\documentstyle[colap,epsf]{article}

\title{Processing Metonymy: a Domain-Model Heuristic Graph Traversal
  Approach {\normalsize\tt (cmp-lg/9604016)}\thanks{This work has been
    partly supported by the European project MENELAS (AIM 2023).}}

\author{Jacques Bouaud, Bruno Bachimont, Pierre Zweigenbaum\\
  DIAM: SIM/AP-HP \& Dept de Biomath\'ematiques, Universit\'e Paris 6\\
  91, boulevard de l'H\^opital F-75634 Paris Cedex 13\\
 \{jb, bb, pz\}@biomath.jussieu.fr}

\newcounter{excnt}
\def\theexcnt{(\arabic{excnt})}

\newif\ifnumberedexample
\numberedexampletrue

\newenvironment{example}{\list{}{}%
\refstepcounter{excnt}
\item[{\rm\ifnumberedexample\theexcnt\else\fi}]%
\sl}{\endlist}

\def\nextexample{\refstepcounter{excnt}
\item[{\em\ifnumberedexample\theexcnt\else\fi}]%
\sl}

\newcommand{\eg}{{\em e.g.}}
\newcommand{\ie}{{\em i.e.}}
\newcommand{\vs}{{\em vs}}

\newenvironment{compact}{\scriptsize\vspace*{-.3ex}\begin{center}}%
{\end{center}\vspace*{-.3ex}\normalsize}

\newcommand{\conc}[1]{[#1]}
\newcommand{\lrel}[1]{$\leftarrow$(#1)$\leftarrow$}
\newcommand{\lrelb}[1]{(#1)$\leftarrow$}
\newcommand{\rrel}[1]{$\rightarrow$(#1)$\rightarrow$}
\newcommand{\rrelb}[1]{(#1)$\rightarrow$}

\newcommand{\rw}[1]{{\bf #1}}

\newcommand{\SetFigFont}[3]{}

\begin{document}

\maketitle
\vspace{-0.5in}
\begin{abstract}
  We address here the treatment of metonymic expressions from a knowledge
  representation perspective, that is, in the context of a text
  understanding system which aims to build a conceptual representation from
  texts according to a domain model expressed in a knowledge representation
  formalism.
  We focus in this paper on the part of the semantic analyser which deals
  with semantic composition.  We explain how we use the domain model to
  handle metonymy dynamically, and more generally, to underlie semantic
  composition, using the knowledge descriptions attached to each concept of
  our ontology as a kind of concept-level, multiple-role qualia structure.
  We rely for this on a heuristic path search algorithm that exploits
  the graphic aspects of the conceptual graphs formalism.
  The methods described have been implemented and applied on French
  texts in the medical domain.  
\end{abstract}

\section{Introduction}
\label{sec:intro}

Under the compositional assumption, semantic analysis relies on the
combination of the meaning representations of parts to build the
meaning representations of a whole.  However, this composition often
needs to call on implicit knowledge which helps to link the two
meaning representations.  This is the case, for instance, in metonymic
expressions, where a word is used to express a notion closely related
to its central meaning.  A well-known stream of work addressing this
phenomenon is the Generative Lexicon theory \cite{PustejovskyCL91}.
At the heart of this theory is a lexical semantic representation
called ``qualia structure''.  Metonymies are considered to correspond
to changes in the semantic types of the words involved, and the qualia
structure provides the basis for performing type coercion in a
generative way.

We address here the treatment of metonymic expressions from a knowledge
representation perspective, in the context of the {\sc Menelas} medical text
understanding system \cite{Zweigenbaum:SCAMC95}.  One of the goals of the
overall system is to assign standardised, medical nomenclature codes to the
input texts (patient discharge summaries).  Semantic analysis starts from a
syntactic representation of each sentence and produces a conceptual
representation.  It is then used by several language-independent,
knowledge-based components to perform inferences (pragmatic enrichment) and
then code assignment \cite{Delamarre:MIM95}. Therefore, the conceptual
representation output by the semantic analyser must be normalised: it must
conform to a knowledge representation canon in which the target nomenclature
codes can be mapped.  The specification of this canon relies on the
description of a rich model of the domain in a knowledge representation
formalism, here Conceptual Graphs (CG) \cite{SowaBOOK84}.

We focus in this paper on the part of the semantic analyser that deals with
semantic composition.  The conceptual representation built must be
abstracted from initial linguistic variation, metonymy being a typical
problem to be addressed.  We explain how we use the domain model to handle
metonymy, and more generally, to underlie semantic composition, using the
knowledge descriptions attached to each concept of our ontology as a kind of
concept-level, multiple-role qualia structure.  The methods described have
been implemented and applied to French texts.

We first recall the problem addressed (section~\ref{sec:metonymy}).  Then,
the proposed method is described (section~\ref{sec:method}) and illustrated
on an example.  We give some information on the implementation and the
results of the analyser (section~\ref{sec:implement}), and discuss the
relative merits of the method (section~\ref{sec:discuss}).

\section{Metonymy and type coercion}
\label{sec:metonymy}

A classical example of metonymy \cite[p.~428ff]{PustejovskyCL91} is
\begin{example}\label{ex:novel}
John began a novel.
\end{example}
where predicate `began' expects an {\em event\/} as its second
argument, so that some way must be found to relate the {\em object\/}
`novel' to an event such as  `to read a novel' or `to write a
novel'.  In our domain (coronary diseases), one often finds expressions
such as

\begin{example}\label{ex:start}
\label{ex:angio-segment}
une angioplastie du segment II
(an angioplasty of segment II)

\nextexample\label{ex:angio-rca}
une angioplastie d'une art\`ere coronaire
(an angioplasty of a coronary artery)

\nextexample\label{ex:angio-hum}
l'angioplastie de Monsieur X
(the angioplasty of Mr X)

\nextexample\label{ex:angio-stenosis}
une angioplastie de la st\'enose
(an angioplasty of the stenosis)

\label{ex:end}
\end{example}
where `angioplasty' is an action performed on a {\em segment\/} of an {\em
  artery\/} to enlarge its diameter, while `stenosis' is the {\em state\/}
of an artery which has a reduced diameter.  These four phrases involve the
object (or ``theme'') of action `angioplasty', \ie, what the angioplasty
operates upon. If one considers that this theme must be a {\em physical
  object}, then examples~\ref{ex:start}--\ref{ex:angio-hum} conform to the
selectional restrictions of `angioplasty', while \ref{ex:angio-stenosis}
violates them.  The mechanism of type coercion \cite{PustejovskyCL91}
consists in converting a word type into another so that semantic composition
can work properly.  \ref{ex:angio-stenosis} is then handled as a metonymy,
where the stenosis and the stenosed object enter a state/thing alternation:
`stenosis' is turned into an `object'.

However, it appears that this phenomenon is dependent on the underlying
types (or ``sorts'') under consideration.  For instance in our ontology,
`segment', `artery', `stenosis' and `human' have four different types, and
are not comparable by the {\sc is-a} relation, \eg\ nothing can be both a
segment and an artery.\footnote{{\tt Segment}, in our ontology, corresponds
  to a portion of space, not of matter.} This is a voluntary, methodological
choice \cite{Bouaud:IJCAIW95}, motivated by the fact that these objects give
rise to different inferences and must not be confused by the reasoning
component.  Additionally, in the target normalised conceptual
representation, what constitutes the specific theme (in our conceptual
model, the {\tt purported\_obj}) of action `angioplasty' must be precisely
defined.  In the context of our application, `angioplasty' acts on an {\tt
  artery\_segment}, a physical object corresponding to a part of an artery,
which happens not to be comparable to any of the four preceding themes of
`angioplasty'.\footnote{Notice, though, that these types are strongly linked
  (by relations other than {\sc is-a}) through the knowledge base models.
  The semantic analyser precisely recovers these links thanks to the
  mechanism presented in this paper.}  Therefore, all four
examples~\ref{ex:start}--\ref{ex:end} must be considered as metonymies.

To handle metonymy, \newcite{FassCOLING88} proposes a method based on a list
of alternations implemented as specific metonymy rules:
Part\-\_for\-\_Whole, Container\-\_for\-\_Contents, etc.
\newcite{SowaINBOOK92} considers metonymies around the term ``Prix
Gon\-court'', originally introduced by \newcite{KayserCOMPINT88}: this term
undergoes different meaning shifts in each of seven example sentences,
ranging from the author who won the prize to the amount of money received.
Sowa discusses how background knowledge could help to process these
metonymies, based on a knowledge description of what ``Prix Goncourt''
involves.

In our system, the target conceptual representation is defined by a domain
model expressed with CGs.  This same model constitutes the resource which
enables the analyser to handle metonymies.  We explain below how results
similar to Pustejovsky's type coercion may be obtained with a method based
on this domain model instead of a qualia structure.

\section{Method}
\label{sec:method}

\subsection{Rationale}

The input to the semantic analyser is the syntactic representation of a
sentence produced by a previous large coverage syntactic analyser
\cite{Berard:AIMIF89}.  This representation connects words, or predicates,
with grammatical relations such as subject, object, oblique object, modifier,
etc.  The output of the semantic analyser is a conceptual graph on which
pragmatic inferences are performed to enrich the representation.

In the semantic lexicon, each word points to one or more conceptual
representations.  The grammatical link between two words in a sentence
expresses a conceptual link between their two associated conceptual
counterparts. The task of the semantic analyser is to identify this
conceptual link.  Rather than including the knowledge needed for this task
in the semantic lexicon, or in a specific rule base, the program will
examine the domain knowledge to resolve the link.  The method relies on a
heuristic path search algorithm that exploits the graphic aspects of the
conceptual graphs formalism.

\subsection{Domain knowledge}

\begin{figure*}[htbp]
\begin{compact}
\setlength{\epsfxsize}{.8\textwidth}
\mbox{\epsffile{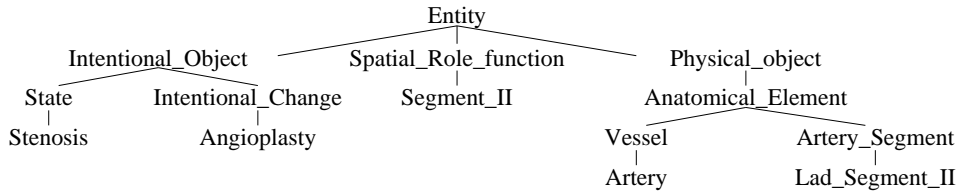}}
\end{compact}
\caption{\label{fig:cth}An extract of the domain ontology.}
\end{figure*}

The main domain knowledge elements consist of the domain ontology
(Fig.~\ref{fig:cth}) which is a subsumption hierarchy of concept types
(henceforth simply `types') and of relation types, and of a set of reference
models attached to the main types.

The reference model of a type represents knowledge about this type as a
conceptual graph (Fig.~\ref{fig:angio}). Basically, a conceptual graph is a
bipartite graph with concept nodes (or concepts) labeled with a type plus an
optional referent, and relation nodes labeled with relation types
\cite{Chein:RIA92}. A model of a given type has an identified head concept
with the same type, and the network of its related concepts represents its
associated knowledge.  Since types are organised in an {\sc is-a} hierarchy,
this knowledge is also inherited.

\begin{figure}[htbp]
\begin{compact}{
\begin{tabbing}
~~ \= ~~~ \= ~~~ \= ~ \= ~~ \= \kill
\rw{Model} Angioplasty(*x) \rw{is} \+ \\
    \conc{Angioplasty: *x}- \+ \\
        \rrelb{pat}\conc{Human\_Being:*pat}\rrel{cultural\_function}\+\+\\
        \conc{Medical\_Subfunction}\rrel{cultural\_role}\conc{Patient} \-\- \\
        \rrelb{agt}\conc{Human\_Being:*doc}\rrel{cultural\_function}\+\+ \\
        \conc{Medical\_Subfunction}\rrel{cultural\_role}\conc{Physician}\-\- \\
        \rrelb{motive}\conc{State\_Of\_Mind}- \+ \\
                \rrelb{state\_of}\conc{Human\_Being:*doc} \\
                \rrelb{content}\conc{Stenosis:*st1} \% \- \\
        \rrelb{purported\_obj}\conc{Artery\_Segment:*as}- \+ \\
                \lrelb{involves}\conc{Stenosis:*st1} \\
                \lrelb{involves}\conc{Internal\_State:*is3} \\
                \lrelb{part}\conc{Human\_Being:*pat} \% \- \\
        \lrelb{descriptive\_goal}\conc{Internal\_State:*is3}- \\
        ...
\end{tabbing}}
\end{compact}
\caption{\label{fig:angio}An extract of reference model for type {\tt
    Angioplasty}.}
\end{figure}

\subsection{Semantic lexicon}

The semantic analyser relies on a two-tier semantic lexicon: one for
predicates, the other for grammatical relations.  Predicates map to
conceptual graphs; most of them are reduced to one concept, since most of
the words in the lexicon are technical terms for which a type exists.
Figure~\ref{fig:semlex} reports some lexical entries.

It is difficult to map grammatical relations to static, predefined
conceptual representations, since their meaning in the domain depends on
their context of use, and mostly on the predicates they link. Besides, one
cannot think of envisioning all the possible uses of such a relation, partly
because of the use of metonymy. The conceptual representation of an actual
grammatical link will therefore be computed dynamically by the semantic
analyser using its context: the linked predicates and domain knowledge.
However, each grammatical relation may have conceptual preferences for types
or for conceptual relations. These preferences are associated with the
grammatical relation.  Our grammatical relations include oblique
complements, so that prepositions in our semantic lexicon are expressed
under this second paradigm (Fig.~\ref{fig:semlex}).

\begin{figure}[htbp]
\begin{compact}{
\begin{tabbing}
~~ \= ~~~ \= ~~~ \= ~ \= ~~ \= \kill
\rw{Entry} angioplastie\_f \rw{is} \conc{Angioplasty: *x}\rw{.}\\
\rw{Entry} stenose\_f \rw{is} \conc{Stenosis: *x}\rw{.}\\
\rw{Entry} segment\_II\_f  \rw{is}\+ \\
 \conc{Segment\_II:*x}-\+ \\
  \rrelb{relative\_to}\conc{Artery} \\
  \lrelb{spatial\_role}\conc{Spatial\_Object} \+\\
    \rrel{zone\_of}\conc{Artery\_Segment}\rw{.}\-\-\- \\
... \\
\rw{Grammatical\_rel} de\_f \rw{:prefers} \+\\
 purported\_obj involved\_obj pat \\
 motivated\_by before\_state after\_state rel\rw{.}\-\\
...
\end{tabbing}}
\end{compact}
\caption{\label{fig:semlex}Some semantic lexicon entries for 
  predicates and a grammatical relation.}
\end{figure}

\subsection{Algorithm}

Given an input triple predicate, grammatical relation, predicate
($P_1;Gr;P_2$), the semantic analyser first replaces the two predicates with
their semantic entries --- two conceptual graphs.  It then endeavours to
link them, that is, to find a concept-level relation between their two head
concepts $C_1$ and $C_2$ that, first, is compatible with the semantic
preferences of grammatical relation $Gr$, and, second, conforms to the
representational canon made of the reference models.

\subsubsection{Design principle.}

The basic idea is to project the two head concepts onto the domain
knowledge and find a plausible concept-level relation between the two.
We implement this by heuristic graph traversal through the reference
models and the type hierarchy, looking for a chain made of concepts
and conceptual relations (\ie\ a linear conceptual graph), which could
link concepts of the same types as $C_1$ and $C_2$ and at the same
time would satisfy the conceptual preferences of $Gr$. Semantic
analysis then consists in solving recursively every grammatical link
starting from the sentence head predicate and then joining the
obtained conceptual chains to build the conceptual representation of
the whole sentence. We focus here only on the link resolution
algorithm.

\subsubsection{Chain production methods.}
\label{sec:chainprod}

We consider that each predicate $P_i$ is associated with the head
concept $C_i$ of a model $M_i$.  Let $T_i$ be the type of $C_i$. We
also assume a partial order on types.  We focus here only on the
strategy for producing the set of all possible chains between $C_1$
and $C_2$.  We can use three methods of increasing complexity to find
chains to link $C_1$ and $C_2$:
\begin{enumerate}
\item \underline{Concept fusion:} the two concepts may be redundant.

If $T_1 \leq T_2$ or $T_1 > T_2$, then $C_1$ and $C_2$ could be merged, and
an empty chain is returned.

\item \underline{Concept inclusion}: a concept may be ``included'' in the
other's model.
\begin{enumerate}
\item For every concept $C'$ of type $T'$ in $M_1$ such that $T' \geq T_2$,
every path between $C_1$ and $C'$ in $M_1$ is a returned chain.
\item For every concept $C'$ of type $T'$ in $M_2$ such that $T' \geq T_1$,
every path in $M_2$ between $C'$ and $C_2$ is a returned chain.
\end{enumerate}

\item \underline{Model join:} two arbitrary concepts in the two models could
be joined. 

For every pair of concepts $(C'_1,C'_2)$ where $C'_i$ of type $T'_i$ is in
$M_i$, and such that $T'_1 \leq T'_2$ or $T'_1 > T'_2$, all the paths
$Paths_1$ between $C_1$ and $C'_1$ in $M_1$ and $Paths_2$ between $C'_2$ and
$C_2$ in $M_2$ are produced. Then, for every pair $(p_1,p_2)$ in
$Paths_1\times Paths_2$, the chain made of the two paths where $last(p_1)$
is joined to $first(p_2)$ is returned.
\end{enumerate}
At this point, we are provided with all chains extracted from the pair of
models $(M_1,M_2)$.

\subsubsection{Model identification.}

The models that associate knowledge to a given predicate $P$ can be ranked
according to their level of generality. The most specific model is the
predicate definition in the semantic lexicon.  The next one is the reference
model associated with the type $T$ of the head concept of the definition.
Then, the following models are the reference models inherited along the
ontology through supertypes of $T$. As the type hierarchy is, in our system,
a tree \cite{Bouaud:IJCAIW95}, the models for a predicate are strictly
ordered.
Considering two grammatically linked predicates, the product of their models
constitutes as many model pairs that can be potentially used to look for
possible chains.  Such pairs are structured by a partial order based on the
generality rank of their members.\footnote{A model pair $(m_1,m_2)$ is more
  specific than $(m'_1,m'_2)$ if $max\_rank(m_1,m_2)$ is less than
  $max\_rank(m'_1,m'_2)$, or if equal, $min\_rank(m_1,m_2)$ is less than
  $min\_rank(m'_1,m'_2)$.}

\subsubsection{Heuristic chain selection.}

At this stage, we are provided with all the possibles chains between $P_1$
and $P_2$ extracted from their models.  The remaining problem is to choose
the most appropriate chain to substitute for $Gr$.  After some
experimentation, we chose the following scheme. The best chain is selected
according to five heuristic criteria: (1) satisfiability of $Gr$
preferences; (2) most specific model pair, \ie, the use of most specific
knowledge associated with words is prefered; (3) simplest chain production
method (see~\ref{sec:chainprod}); (4) most specific or highest priority of
$Gr$ preferences; (5) shorter chain length.  When multiple chains remain in
competition, one is selected randomly.

To reduce search, the link resolution strategy does not consider all
possible chains, and implements the first two criteria directly in the chain
production step.  Chains that violate $Gr$ preferences are discarded, and
model pairs are explored starting from the most specific pair.


\subsection{An example}
\label{sec:example}

Let us illustrate the resolution on example~\ref{ex:angio-segment} (an
angioplasty of segment II). The input triple is
(angioplastie\_f;de\_f;segment\_II\_f). The corresponding types, {\tt
  Angioplasty} and {\tt Segment\_II}, are not compatible and the ``fusion''
method fails.  The ``inclusion'' method also fails since no model for
angioplastie\_f includes a concept compatible with {\tt Segment\_II}, and no
model for segment\_ii\_f includes a concept compatible with {\tt
  Angioplasty}.

However, with the ``join'' method, the algorithm identifies 6063
possible chains that satisfy the preferences attached to preposition de\_f
(Fig.~\ref{fig:semlex}). The selected chain uses the reference model of
{\tt Angioplasty} (Fig.~\ref{fig:angio}) and the definition graph for
segment\_II\_f (Fig.~\ref{fig:semlex}) which are connected on concept {\tt
  Artery\_Segment}.  The resulting conceptual
representation joins the two corresponding paths: \\
\begin{compact}
\begin{tabbing}
~~~ \= \kill
\conc{Angioplasty}\rrel{purported\_obj}\conc{Artery\_Segment}. \\
~\\
\conc{Artery\_Segment}\lrel{zone\_of}\conc{Spatial\_Object}\\
     \> \` \rrel{spatial\_role}\conc{Segment\_II}.
\end{tabbing}
\end{compact}
into
\begin{compact}
\begin{tabbing}
~~~~~~~~~~~ \= ~~~~ \= \kill
\conc{Angioplasty}\rrel{purported\_obj}\conc{Artery\_Segment}\\
\>  \lrel{zone\_of}\conc{Spatial\_Object} \\
\> \> \` \rrel{spatial\_role}\conc{Segment\_II}.
\end{tabbing}
\end{compact}

This representation reflects the fact that in the context of an
`angioplasty', `segment II' is considered from the point of view of the
physical artery segment the angioplasty is to act upon (instead of the
spatial notion {\tt Segment\_II} expresses).

\begin{table*}[htbp]
\caption{\label{tab:meto-angio-of}Conceptual representations obtained for
  sentences~(2)--(5).}
\begin{compact}
\newcommand{\resrow}[6]{\hline
#1 #2&#3&#4&#5\\
\multicolumn{4}{l}{#6}}

\newcommand{\resrowb}[2]{
\multicolumn{4}{r}{#2}\\
&&&#1}
~

\begin{tabular}{l@{~~~~~~~~~~~~~~~~~~~~~~~~~~~~~~~~~~}ccc}
\resrow{\em (\#)}{\em phrase}{\em total chains}{\em method}{\em models}{\em partial chains selected}
\\ \hline
\resrow{\ref{ex:angio-segment}}{`angioplasty of segment II'}
{6063}
{join}
{Angioplasty}
{\conc{Angioplasty}\rrel{purported\_obj}\conc{Artery\_Segment}}
\\
\resrowb
{`segment II' definition}
{\conc{Artery\_Segment}\lrel{zone\_of}\conc{Spatial\_Object}\rrel{spatial\_role}\conc{Segment\_II}}
\\
\resrow{\ref{ex:angio-rca}}{`angioplasty of a coronary artery'}
{2387}
{inclusion}
{Angioplasty}
{\conc{Angioplasty}\rrel{purported\_obj}\conc{Artery\_Segment}\lrel{part}\conc{Coronary\_Artery}}
\\
\resrow{\ref{ex:angio-hum}}{`angioplasty of Mr X'}
{3633}
{inclusion}
{Angioplasty}
{\conc{Angioplasty}\rrel{purported\_obj}\conc{Artery\_Segment}\lrel{part}\conc{Human\_Being}}
\\
\resrow{\ref{ex:angio-stenosis}}{`angioplasty of a stenosis'}
{2217}
{inclusion}
{Angioplasty}
{\conc{Angioplasty}\rrel{purported\_obj}\conc{Artery\_Segment}\lrel{involves}\conc{Stenosis}}
\\
\hline
\end{tabular}
\end{compact}
\end{table*}

\section{Implementation and results}
\label{sec:implement}

This analyser has been implemented on top of a conceptual graph processing
package embedded in Common Lisp.  In the current state, the ontology
contains about 1,800 types and 300 relation types; over 500 types have their
own reference model; the lexicon defines over 1,000 predicates and about 150
grammatical relations and prepositions. The analyser correctly handles
typical expressions found in our texts, including
examples~\ref{ex:start}--\ref{ex:end} (see table~\ref{tab:meto-angio-of}).
The complete processing chain has been tested on a set of 37 discharge
summaries (393 sentences, 5,715 words) \cite{Zweigenbaum:SCAMC95}.  This corpus
included development texts, so the results are somewhat optimistic; on the
other hand, the system is in an incomplete state of development.  The test
consisted in code assignment and answering a fixed questionnaire, the gold
standard being given by health care professionals.  Overall recall and
precision were measured at 48~\% and 63~\% on the coding task, and 66~\%
and 77~\% on the questionnaire task.

No evaluation has been performed on more basic components of the system; we
can however provide statistics drawn from the global test for the semantic
analyser.  For 274 sentences received, the link resolution procedure was
called on 8,749 grammatical links and explored 247,877 chains, with an
average of 28 chains per call and 904 per sentence.  The number of paths
found depends heavily on the richness of the models used, which varies with
the types involved.  For instance, the model for type {\tt angioplasty}
(involved in table~\ref{tab:meto-angio-of}) is central in the domain.  It is
the most complex in the knowledge base and contains 54 concepts and 78
relations, which accounts for the greater number of paths found in these
examples.

However, inadequate expansions are sometimes made due to lack of models, or
to their complexity, which makes the heuristic principles not selective
enough.  Such limitations also stem from a lack of ``actual'' semantic
knowledge.  The semantic analyser goes directly from grammatical relations
to conceptual relations without any intermediate semantic representation.
Useful information, such as the argumental or thematic structure of
predicates (\eg, \newcite{Melcuk:BOOK95}, \newcite{Pugeault:COLING94}),
could probably overcome some of its shortcomings.

\section{Discussion}
\label{sec:discuss}

One could compare this approach to a concept-based, multi-role qualia
structure.  The semantic definition of a word is here the reference model of
its head concept type; each relation path starting from the head concept of
this reference model is similar to a qualia role, in that it describes one
of the semantic facets or possible uses of the word.  In the context of a
predicate, one of the concepts in the reference model is selected as the
incoming point of a link from the predicate's meaning representation.

The concept-oriented domain-model approach advocated here hypothesizes that
the behaviour of words is driven by their conceptual roles in the domain.
This has the advantage of factoring knowledge at the conceptual level,
rather than having to distribute it at the level of words.  This knowledge
can then be shared by several words.  Sharing even occurs across
languages (\eg\ Dutch \cite{Spyns:MEDINFO95}).

Moreover, the type hierarchy allows concepts, hence words, to inherit
reference models from more abstract concepts, thus enabling more sharing and
modularity.  The distinction between local information and information
inherited through the hierarchy is furthermore exploited when ranking
different chains between two concept types.  

Another difference resides in the way flexibility is obtained.  In
Pustejovsky's coercion mechanism \cite{PustejovskyCL91}, the argument's
semantic type changes for a semantic type found in one of its qualia.  In a
variant approach \cite{Mineur:95}, a word has no a priori semantic type; it
is selected at composition time among the types found in the qualia.  In our
approach, the head concept type associated with an argument does not change.
The chain found between this concept and the predicate's head concept only
brings forward intermediate concepts and relations which are actualised in
the presence of the predicate, and lead to a particular representation of
their meaning.  As a side-effect, this approach is able to handle sentences
like~\ref{ex:double}--\ref{ex:doublefin}:
\begin{example}\label{ex:double}
John bought a long novel
{\rm\cite{Godard:EACL93}}

\nextexample\label{ex:doublestenose}
an angioplasty of a severe stenosis
\label{ex:doublefin}
\end{example}
Since the modifier (long, severe) and the action (verb `bought', noun
`angioplasty') require incompatible types of the same noun (novel: event
\vs\ object, stenosis: state \vs\ object), type changing via coercion cannot
work on such sentences. This problem does not occur in our approach.

Type coercion assumes that the predicate drives semantic composition,
and that the semantic representation of the argument must adapt to it.
In our method, both predicate and argument can make a step towards
finding their semantic link.  The resulting conceptual chain, as a
whole, represents both the specific facet of the argument which is
involved in the sentence and the conceptual role it plays in the
predicate.

The preferences that grammatical relations assign to conceptual relations
drive path selection, taking into account the specific syntactic context in
which a semantic composition is to occur.  This is crucial to let, \eg,
prepositions, influence the choice of the conceptual link and the resolution
of the metonymy.

\section{Conclusion}
\label{sec:conclusion}

The overall goal of the {\sc Menelas} text understanding system was to build
a normalised conceptual representation of the input text. The aim of
semantic analysis, in this context, is to build a representation which
conforms to a domain model.  We therefore experimented how this domain model
could help semantic analysis to go from the flexibility of natural language
to a constrained conceptual representation, a typical problem encountered
being metonymy.  The approach presented here shows how this can be
performed.  It has been fully implemented, and used with a reasonable size
knowledge base as a part of the {\sc Menelas} text understanding system.

Metonymy processing is based on the domain model.  Provided
a new domain and task, with the corresponding domain model, this enables the
generic method to adapt directly to this new domain and give results that
are specific to it.  Building such a domain model is generally feasible in
sufficiently limited domains, typically, technical domains.
Much of the strength of the method then hinges on the quality of the domain
model: the concept type hierarchy and the attached reference models must be
built in a principled way \cite{Bouaud:IJCAIW95}.

\newcommand{\etal}{{\it et al.}}

\end{document}